%% file: main.tex
\documentclass[10pt,twocolumn,letterpaper]{article}

\usepackage{wacv}
\usepackage{times}
\usepackage{epsfig}
\usepackage{graphicx}
\usepackage{amsmath}
\usepackage{amssymb}
\usepackage{booktabs}
\usepackage[accsupp]{axessibility}

\usepackage[utf8]{inputenc} % allow utf-8 input
\usepackage[T1]{fontenc}    % use 8-bit T1 fonts
\usepackage{url}            % simple URL typesetting
\usepackage{booktabs}       % professional-quality tables
\usepackage{amsfonts}       % blackboard math symbols
\usepackage{nicefrac}       % compact symbols for 1/2, etc.
\usepackage{microtype}      % microtypography
\usepackage{xcolor}         % colors
\usepackage{algpseudocode}
\usepackage{algorithm}
\usepackage{adjustbox}
\usepackage{lipsum}
\usepackage{array}
\usepackage{soul}
\usepackage{bm}
\usepackage{multirow}
\usepackage{capt-of}
\usepackage{wrapfig}
\usepackage{xspace}
\usepackage{subcaption}
\usepackage[size=small]{caption}

\def\wacvPaperID{1220}
\wacvalgorithmstrack
\wacvfinalcopy
\ifwacvfinal
\usepackage[breaklinks=true,bookmarks=false]{hyperref}
\else
\usepackage[pagebackref=true,breaklinks=true,colorlinks,bookmarks=false]{hyperref}
\fi

\input{macros.tex}

\begin{document}

\title{Physically Plausible Animation of Human Upper Body from a Single Image}

\author{Ziyuan Huang*\\ National Taiwan University
\and Zhengping Zhou*\\ Stanford University
\and Yung-Yu Chuang\\ National Taiwan University
\and Jiajun Wu \\ Stanford University
\and C. Karen Liu \\ Stanford University}

\maketitle
\thispagestyle{empty}

\input{0_abstract}

\input{1_introduction}
\input{2_relatedwork}
\input{3_method}

\input{4_experiments}
\input{5_discussion}

\myparagraph{Acknowledgments.} This work is in part supported by the Toyota Research Institute (TRI), the Stanford Institute for Human-Centered AI (HAI), Samsung, and Amazon.

{\small
\bibliographystyle{ieee_fullname}
\bibliography{egbib}
}

\end{document}

%% file: macros.tex
\newcommand{\imgrid}[3]{
        \centering
    \begingroup
    \setlength{\tabcolsep}{0.6pt} 
    \renewcommand{\arraystretch}{0.2} 
    \newcommand{\ph}[1]{
     \framebox(#1, #1){\parbox{#1\unitlength}{\centering ##1}}
     }
     \newcommand{\im}[1]{\includegraphics[width=#1]{##1}}
    \begin{tabular}{#2}
        #3
    \end{tabular}
    \endgroup
}

\hypersetup{
  colorlinks=true,
  linkcolor=blue!50!red,
  urlcolor=green!70!black
}

\newcommand{\x}[1]{\bm{x}^{#1}}

\newcommand{\xGT}[1]{\bar{\bm{x}}^{#1}}
\newcommand{\dx}[1]{\triangle \bm{x}^{#1}}
\newcommand{\dxGT}[1]{\triangle \bar{\bm{x}}^{#1}}
\newcommand{\s}[1]{\bm{s}^{#1}}
\newcommand{\I}[1]{I^{#1}}

\newcommand{\q}{\bm{q}}

\renewcommand{\a}[1]{\bm{a}^{#1}}

\newcommand{\D}{\mathcal{D}}

\newcommand{\g}[1]{\bm{g}^{#1}}
\newcommand{\h}[1]{\bm{h}^{#1}}

\definecolor{darkyellow}{rgb}{0.82, 0.41, 0.12}
\definecolor{darkgreen}{rgb}{0.0, 0.5, 0.0}
\definecolor{darkred}{rgb}{0.8, 0.0, 0.0}
\definecolor{darkblue}{rgb}{0.0, 0.0, 0.53}

\newcommand{\downloads}[1]{\url{http://mocha.stanford.edu/downloads#1} (username/password: {\tt zhengping})}

\algnewcommand{\LineComment}[1]{\State \(\triangleright\) #1}

\newcommand{\figref}[1]{Figure~\ref{fig:#1}}
\newcommand{\tabref}[1]{Table~\ref{tab:#1}}
\newcommand{\secref}[1]{Section~\ref{sec:#1}}
\newcommand{\model}{PUBA\xspace}

\makeatletter
\DeclareRobustCommand\onedot{\futurelet\@let@token\@onedot}
\def\@onedot{\ifx\@let@token.\else.\null\fi\xspace}

\def\eg{e.g\onedot}

 \def\vs{vs\onedot}
 
\def\etal{et al\onedot}
\makeatother

\newcommand{\myparagraph}[1]{\vspace{-12pt}\paragraph{#1}}
\newcommand{\myitem}{\vspace{-3pt}\item}

%% file: 0_abstract.tex
\vspace{-20pt}
\begin{abstract}
\vspace{-10pt}
We present a new method for generating controllable, dynamically responsive, and photorealistic human animations. Given an image of a person, our system allows the user to generate Physically plausible Upper Body Animation (\model) using interaction in the image space, such as dragging their hand to various locations. We formulate a reinforcement learning problem to train a dynamic model that predicts the person’s next 2D state (i.e., keypoints on the image) conditioned on a 3D action (i.e., joint torque), and a policy that outputs optimal actions to control the person to achieve desired goals. The dynamic model leverages the expressiveness of 3D simulation and the visual realism of 2D videos. \model generates 2D keypoint sequences that achieve task goals while being responsive to forceful perturbation. The sequences of keypoints are then translated by a pose-to-image generator to produce the final photorealistic video.
\let\thefootnote\relax\footnotetext{* indicates equal contribution.}
\end{abstract}

%% file: 1_introduction.tex
\vspace{-20pt}
\section{Introduction}
\vspace{-5pt}
Physics-based 3D character animation provides powerful tools to create controllable and interactable human agents using physical forces. These techniques can produce entirely novel movements grounded by laws of physics without any training data, but the visual appearance of the agent is still not on par with photorealistic images. In contrast, data-driven image-based motion synthesis methods are effective in synthesizing photorealistic videos, but they are not capable of creating novel movements based on unpremeditated physical interaction with the agent.

\begin{figure*}
\begin{center}
    \centering\small
    % \captionsetup{type=figure}
    \imgrid{75pt}{cccccc}{
       \im{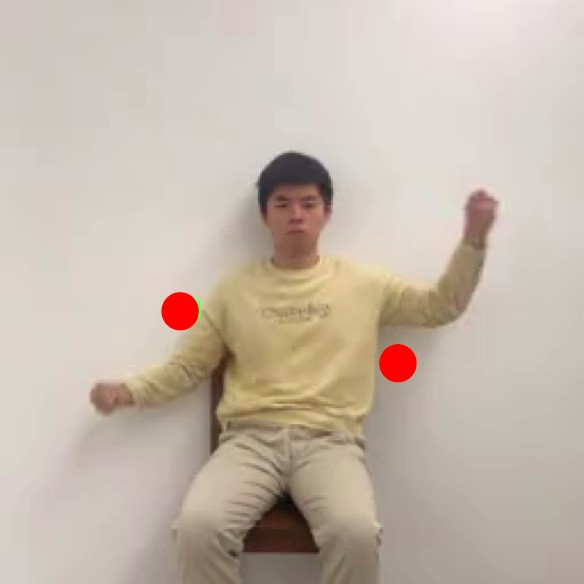} & 
       \im{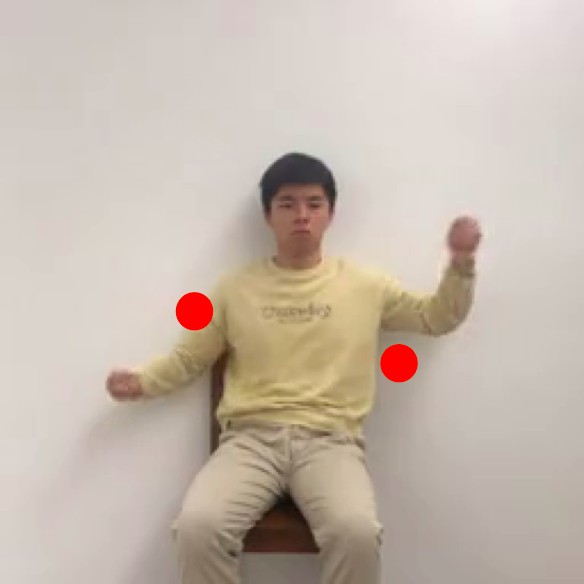} &
       \im{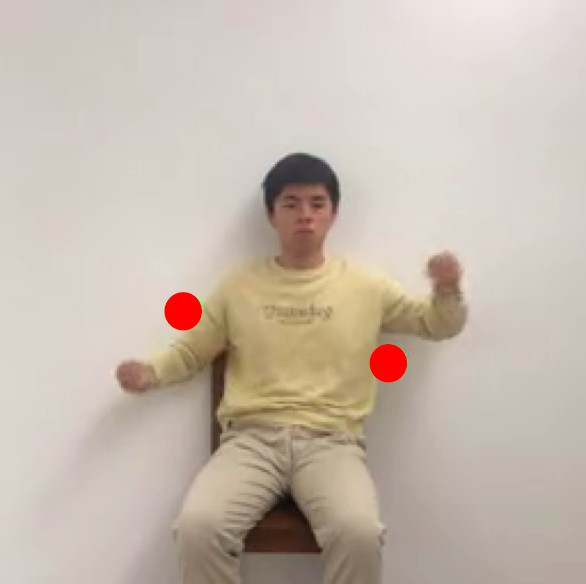} &
       \im{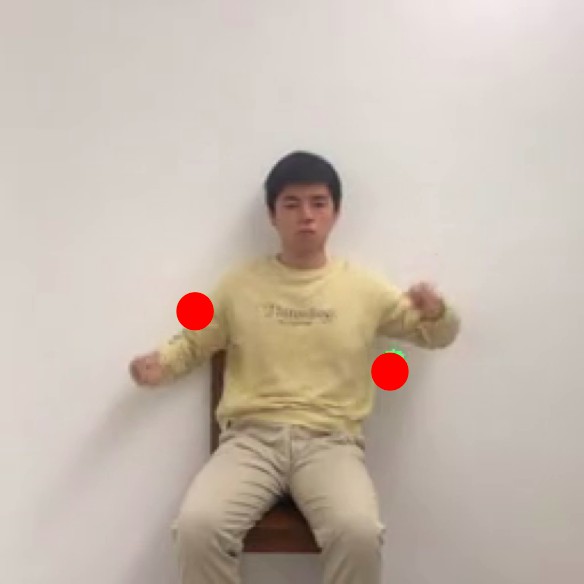} &
       \im{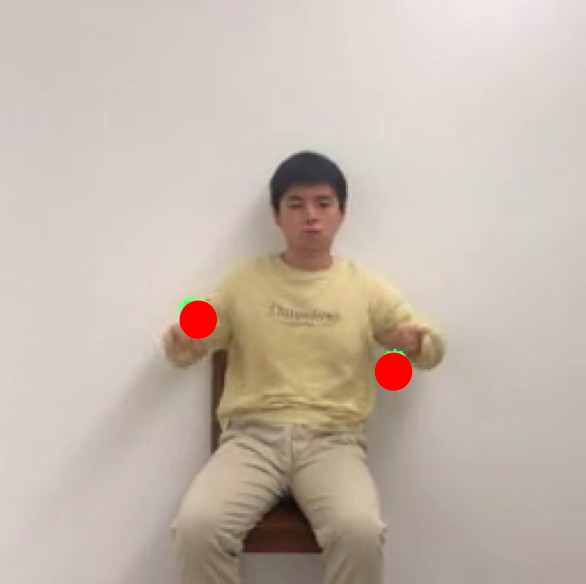} &
       \im{assets/teaser/000004_A.jpg}
       \\
       \im{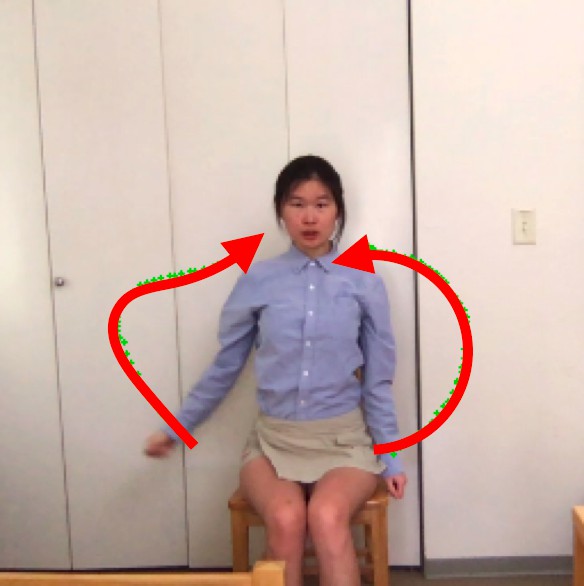} & 
       \im{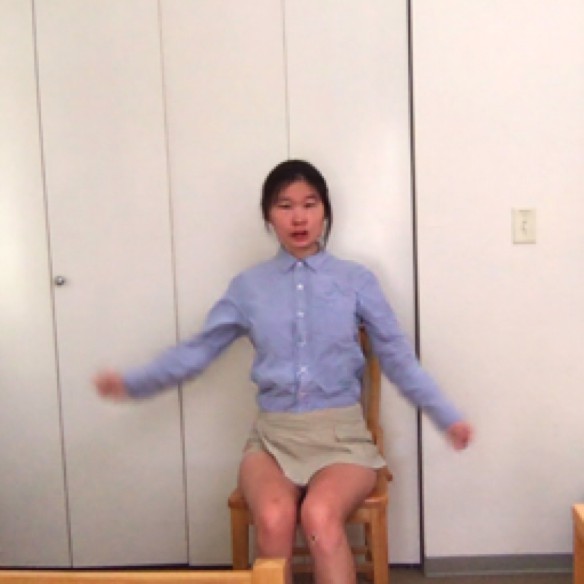} &
       \im{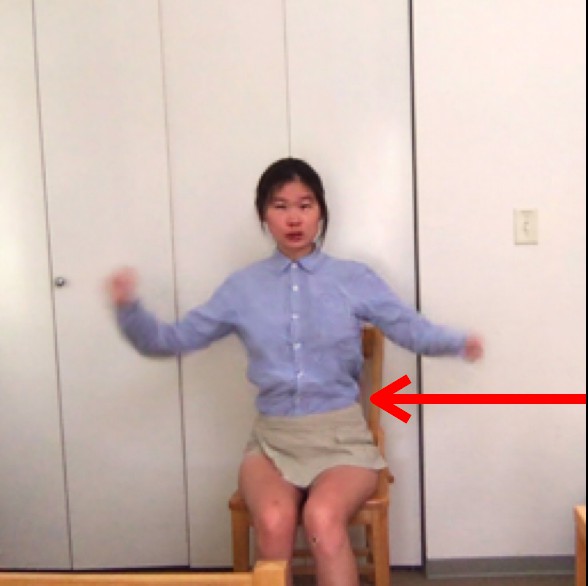} &
       \im{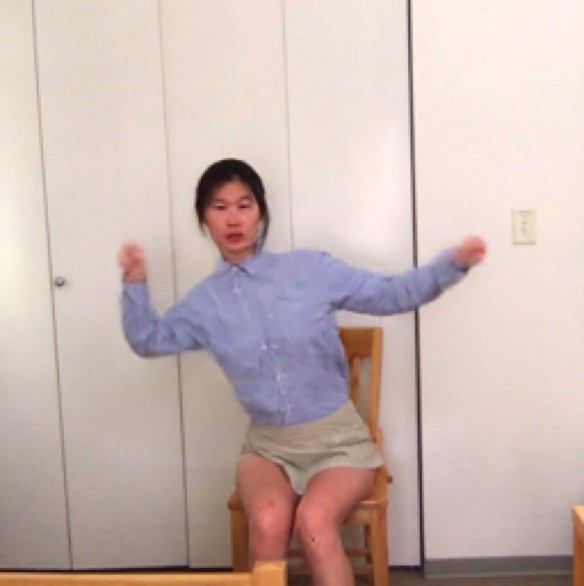} &
       \im{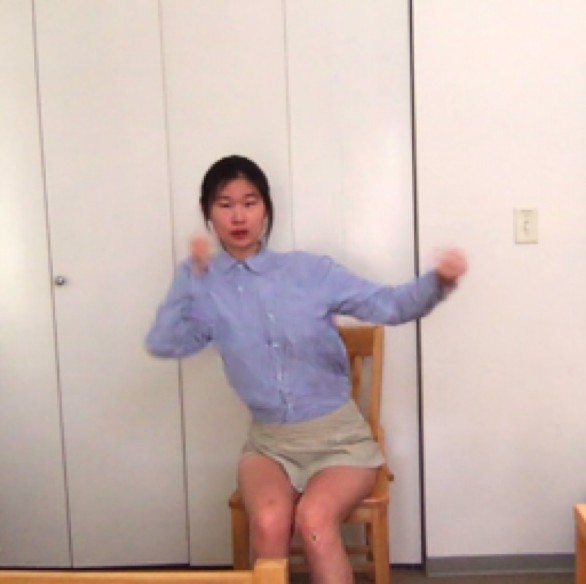} &
       \im{assets/teaser/000004_B.jpg}
       \\
       \\
       Input image & $t=10$ & $t=20$ & $t=30$ & $t=40$ & $t=50$ 
       \\
    }
   \vspace{-5pt}
    \caption{Given an image with a person (first column), our system allows the user to animate the person using intuitive 2D interaction in the image space. Top: The user specifies two target locations for the wrists (shown as red dots). Bottom: The user specifies desired trajectories for hands to track (shown as red curves). At $t=20$, the user creates a force vector that pushes the person to the left (shown as the red arrow).
   }
   \vspace{-20pt}
    \label{fig:teaser}
\end{center}
\end{figure*}

Can we combine the expressive power that comes with physics-based 3D animation with the rich appearance offered by 2D videos, such that the agent is physically controllable, interactable, and photorealistic? We propose a new approach that splits the difference between the 2D and 3D representations---our dynamic model operates on a \emph{2D state} and a \emph{3D action}. We represent states as 2D keypoints to avoid conversion between 3D and 2D representations, because our final products are 2D images. On the other hand, we choose to represent actions as 3D joint torques, because they contain more information to respect human kinematic and dynamic constraints. Using this hybrid representation, we train policies to control the person's upper body in the photo to achieve desired tasks. Combining the policy with the dynamic model, our system can animate the person by dragging her hands to arbitrary 2D locations or pushing the person with an arbitrary force. The person in the photo will follow the user's commands or react to physical forces in a human-like and physically plausible way (Figure \ref{fig:teaser}). 

Our system, named Physically plausible Upper Body Animation (\model), consists of two modules: motion synthesis and photorealistic rendering. We cast motion synthesis into a reinforcement learning problem with the hybrid representation of 2D states and 3D actions. Specifically, we define a 2D state as a set of keypoints on the person in the image and a 3D action as a torque vector applied to the person's joints. We first train a transition function that predicts the next 2D keypoints from the current 2D keypoints under the effect of 3D torque vectors. Subsequently, we train a policy that takes as input 2D keypoints and outputs a 3D torque vector to maximize the long-term reward that defines the task. Once the policy and the transition function are trained, we can generate sequences of 2D keypoints that achieve the task while being responsive to forceful perturbation applied by the user. The keypoint sequences are then translated into photorealistic images via a pose-to-image generator \cite{chan2019everybody}. 

\model has two key advantages: First, the motion synthesis module and the photorealistic rendering module are decoupled and can be trained separately, allowing it to learn from unpaired 3D mocap data and 2D images; Second, the policy and dynamics models operate fully on 2D keypoints, bypassing the need of monocular 3D pose reconstruction. Our system design avoids the failure case when the predicted 3D pose cannot overlay with the input image and removes the dependency on a complex physics-based animation pipeline.

%% file: 2_relatedwork.tex
\vspace{-2pt}
\section{Related Works}
\vspace{-3pt}

\paragraph{Image animation.}
A body of research aims to create an animation for a class of or a specific object in a source image by following the movements of a driving video, based on  cycle consistency~\cite{recycle-gan-http://link.springer.com/10.1007/978-3-030-01228-1_8}, estimation of optical flow~\cite{siarohin2019animating}, or affine transformation~\cite{siarohin2019first} of unsupervisedly detected 2D keypoints. A special line of work focuses on human body pose transfer, where the central goal is to synthesize a person in novel poses. Representative works include 2D methods that use direct mapping from
pose to image~\cite{chan2019everybody}, estimated body part spatial transformations~\cite{balakrishnan2018synthesizing, https://renyurui.github.io/GFLA-web/}, attention maps~\cite{https://arxiv.org/pdf/1904.03349.pdf}, modal bases~\cite{davis2015image}, and disentanglement of color and pose~\cite{https://compvis.github.io/vunet}, as well as 3D methods that take advantages of dense pose~\cite{https://arxiv.org/abs/1809.01995},  reconstruction of parametric body models~\cite{https://svip-lab.github.io/project/impersonator}, or
person-specific textured 3D character models~\cite{liu2019neural}. 

A key limitation of those methods is that complete reference motion must be provided.
In contrast, our method allows the user to generate animation by  specifying a concise goal, for instance ``track this mouse trajectory'' or ``reach this target location with your wrists''. The synthesized motion can be further edited by physical forces, thus generating physically plausible behaviors such as the person ``struggles to prevent falling when suddenly pushed to the side''. 

\myparagraph{Motion prediction.} 
Opposite to works that synthesize motions based on a reference motion, another line of research focuses on predicting the motion solely based on past or future observations, where the predicted motion can either be represented as RGB videos \cite{struct-v-rnn-http://arxiv.org/abs/1906.07889}, or as 3D body mesh sequences \cite{http://arxiv.org/abs/1908.04781, humanMotionKanazawa19} based on the prior art of parametric body surface models \cite{loper2015smpl} and monocular 3D body pose and shape reconstruction \cite{bogo2016keep, kanazawa2018end, kolotouros2019learning,shimada2020physcap,yuan2021simpoe,xie2021physics}. For future prediction, an autoregressive model is typically employed that takes the output from the last step as input, such as Struct V-RNN~\cite{struct-v-rnn-http://arxiv.org/abs/1906.07889} or a casual model~\cite{http://arxiv.org/abs/1908.04781}. The autoregressive model has also proved to be effective for synthesizing long-term 3D motion from scratch, such as the acLSTM model~\cite{li2017auto}.

Since the synthesized motion is based on past predictions or generated purely from scratch, they are barely controllable and are only applicable to highly specific and predictable motions, such as sports or martial arts. However, we borrow the idea of autoregressive prediction for our dynamics model, which is controlled by a per-step torque input.

\myparagraph{Physical simulation-based character animation.}
On the spectrum of ``motion controllability'', physically simulated character animation stands on ``the most controllable'' extreme, while motion prediction stands on the other. For synthesizing motions that react realistically to perturbations, controllers are widely used for moderating the character's movements in the 3D environment. To bring a similar level of controllability and physical realism into the 2D pixel space, we borrow the idea from the deep reinforcement learning (DRL) based animation systems \cite{peng2018deepmimic, peng2018sfv} that learn policies for user-specified goals. The key difference is that our policy takes 2D keypoint-based observations, while they operate fully in 3D and require joint angle-based observations, which are usually difficult to infer from a single view. 

%% file: 3_method.tex
\vspace{-13pt}
\section{Method}
\vspace{-3pt}

Given a person in a photo $\I{0}$, our \model system enables the user to animate the person using intuitive 2D interaction, such as dragging or pushing at different parts of the person, and create a photorealistic video $\I{1:T}$. We view the person in the photo as a reinforcement learning agent operating in a hybrid representation of 2D state space and 3D action space. The task of the agent is to achieve the target locations specified by the user and respond dynamically to perturbation applied by the user.

\begin{figure}[t]
\centering
\includegraphics[width=.4\textwidth]{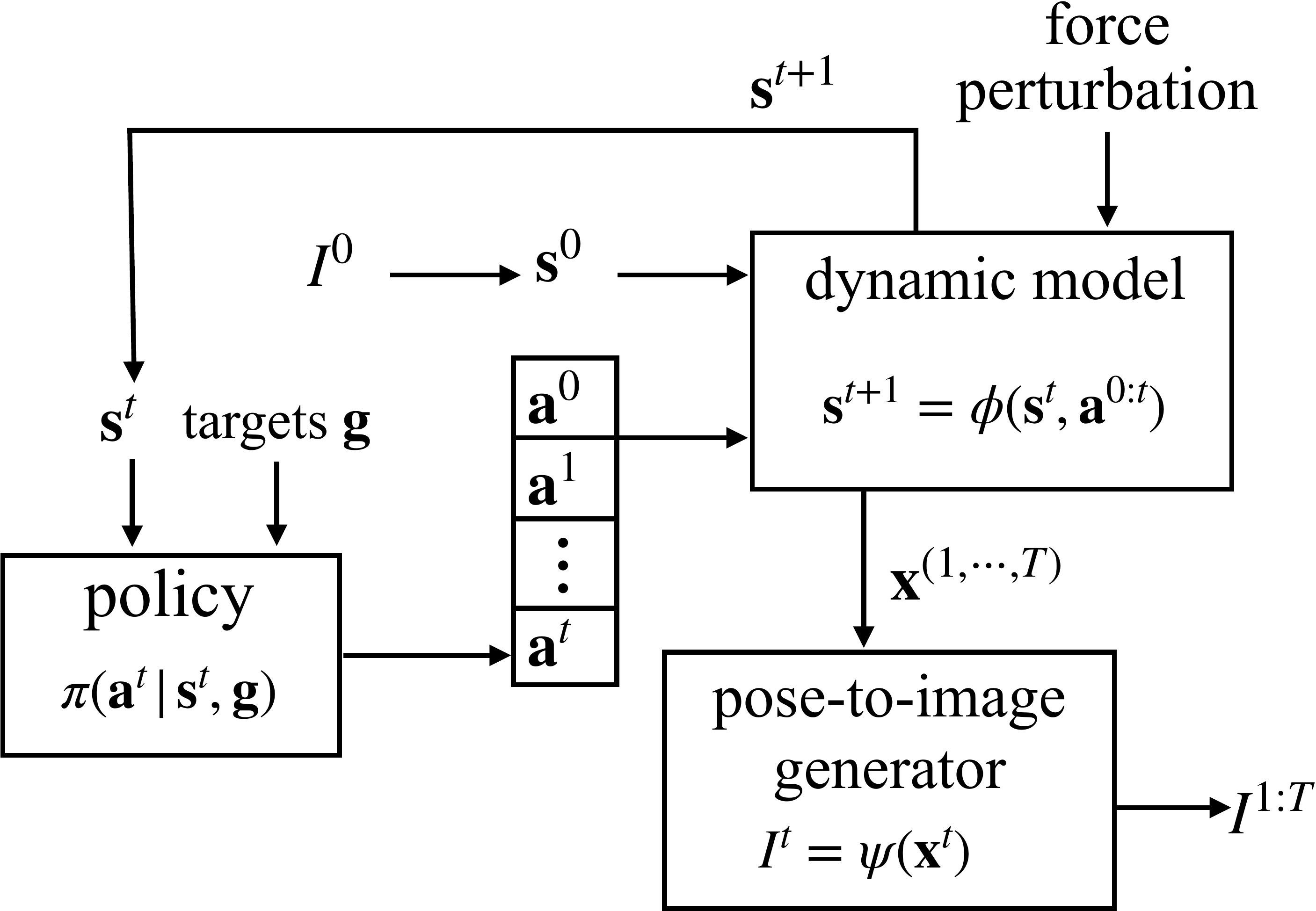}
\vspace{-3pt}
\caption{Overview of our system.}
\vspace{-10pt}
\label{fig:tldr}
\end{figure}

Our system consists of three components (Figure \ref{fig:tldr}):
\begin{itemize}
    \vspace{-5pt}
    \item a dynamic model $\s{t+1} = \phi(\s{0}, \a{0:t})$,
    \vspace{-5pt}
    \item a goal-conditioned policy $\a{t} \sim \pi(\cdot | \s{t}, \bm{g})$, and
    \vspace{-5pt}
    \item a pose-to-image generator $\I{t} = \psi(\x{t})$.
\end{itemize}    
A 2D pose, $\bm{x} \in \mathbb{R}^{2n}$ is defined by $n$ landmark points on the person's body, such as shoulders and elbows. A state $\bm{s} \in \mathbb{R}^{4n}$ is a concatenation of a 2D pose and its offset from the 2D pose at the previous time step, $\s{t} = (\x{t}, \dx{t})$. An action is defined as a 3D torque vector, $\bm{a} \in \mathbb{R}^m$, where $m$ is the number of actuated degrees of freedom (DOFs) of the proxy 3D agent. Finally, our policy is also conditioned on the 2D target locations, $\bm{g}$, for the specified body parts.

\subsection{Dynamic Model} 

\label{sec:fd}

\begin{figure}[t]
    \centering
        \includegraphics[width=0.9\linewidth]{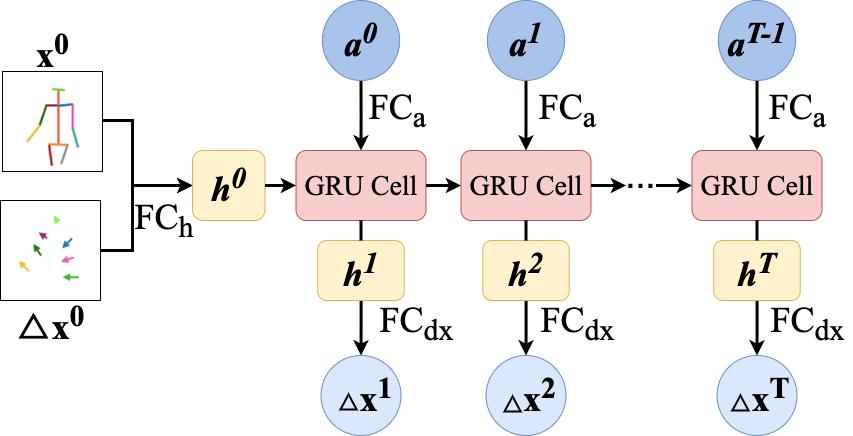}
        \vspace{-5pt}        
        \caption{The dynamics model. Starting from initial 2D keypoint coordinates $\x{0}$ and offsets $\dx{0}$, the GRU model takes one torque $\a{t}$ at each step $t$ as input, and predicts the per-step 2D keypoint offsets $\dx{t}$. $FC_h$, $FC_a$, $FC_{dx}$ are fully connected layers with $\tanh$ activation.}
        \vspace{-10pt}        
        \label{fig:rnn}
\end{figure}

The dynamic model is trained to simulate a sequence of states $\s{1 :t+1}$, given the initial state $\s{0}$ and a control sequence $\a{0:t}$ (Figure \ref{fig:rnn}). Specifically, at time $t$, the network predicts a sequence of $\dx{1:t+1}$ and integrates them to get the 2D pose at $t+1$ as $\x{t+1} = \x{0}+\sum_{t'=1}^{t+1} \dx{t'}$, which is then concatenated with $\dx{t+1}$ to compose the next state $\s{t+1} = (\x{t+1}, \dx{t+1})$.

\begin{figure}[t]
    \centering
        \includegraphics[width=0.9\linewidth]{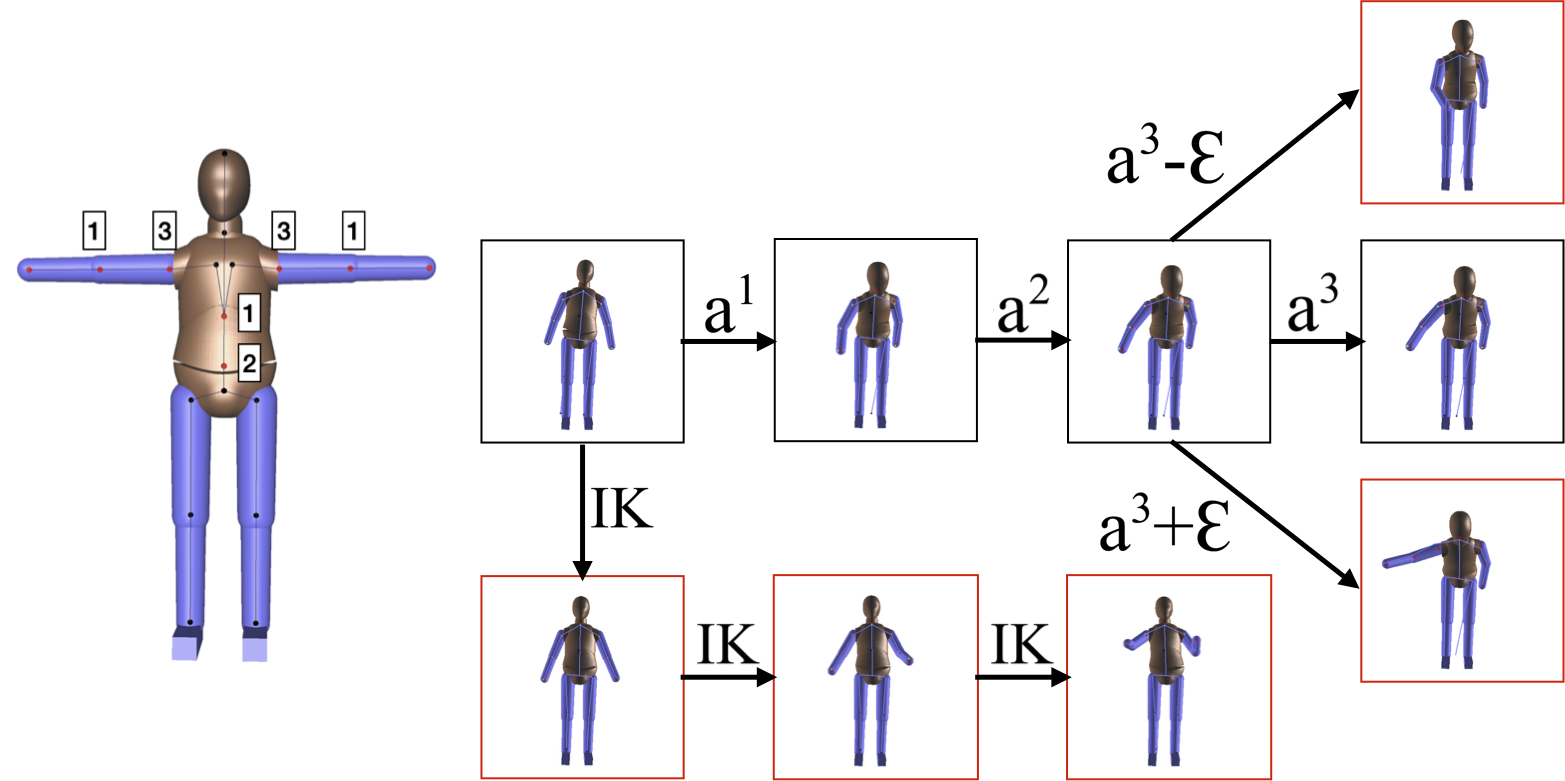}
        \vspace{-5pt}
        \caption{Left: The location and DOFs for each joint on our 3D human agent. Right: The training data for the dynamic model. A sample motion sequence from the mocap dataset is shown in black boxes. 
        The augmented poses are shown in red boxes.} 
        
        \vspace{-12pt}
        \label{fig:training_data}
\end{figure}

\myparagraph{Training data.}
Training the dynamic model $\phi$ requires a large set of training data that covers the joint space of state and action. We use motion capture dataset $\mathcal{D}$ from Human 3.6M \cite{ionescu2013human3} and augment it with a set of synthetic human motions $\mathcal{D}'$. We build a 3D human agent represented as an articulated rigid body system to process the mocap data (Figure \ref{fig:training_data} Left). We first apply inverse kinematics to obtain joint configurations $\{\q^{1:T}\}$ from the raw mocap data, where $\bm{q} \in \mathbb{R}^m$ contains the actuated degrees of freedom of the human agent. To derive the joint torques $\{\a{0:T}\}$, we compute inverse dynamics from trajectories of $\bm{q}$, as well as the trajectories of $\dot{\bm{q}}$ and $\ddot{\bm{q}}$ approximated by finite differencing. To obtain the 2D keypoints $\{\xGT{0:T}\}$, we simply compute the 3D landmark points via forward kinematics and project them to the 2D image space.

We use two different approaches to create augmented human motion dataset $\mathcal{D}'$ (\figref{training_data}). First, since the policy training process usually involves adding exploration noise to actions for improving the robustness of the policy, we augment the training data by perturbing the per-step torque $\{\a{0:T}\}$, computed by inverse dynamics, with noise sampled from a zero-centered Gaussian distribution, and then forward simulate to get a new state. The standard deviation of the Gaussian matches half of the standard deviation of the joint torques computed from the mocap dataset. Second, we randomly generate 3D points within the human range of motion as the target points for the specified body parts (\eg, the wrist), and use inverse kinematics to generate additional reaching motion for training. Using these two methods for data augmentation, we can generate as much training data as we need. We augment 330,000 frames into the original dataset, which contains 141,021 frames. The dynamic model trained with such an augmented dataset performs more reliably when being queried by the policy during training, as shown quantitatively in various scenarios in \secref{exp}. 

\myparagraph{Model architecture.}
We use a gated recurrent unit (GRU)~\cite{gru} network to model the temporal dependency in human motion. The hidden state, which implicitly models the evolving kinematic state, is initialized through a fully connected layer that takes $\s{0} = (\x{0}, \dx{0})$ as input, and is updated at each step according to the input torque $\a{t}$. We
denote the matrix multiplication operator as $\circ$, the vector concatenation operator as $[\cdot]$, and the GRU cell update rule as $\text{GRU}(\cdot)$:
\begin{equation}
        \h{t} = \begin{cases}
                \tanh (W_h\circ[\x{0}, \dx{0}]+\bm{b}_h), & t = 0;\\
                \text{GRU}(\h{t-1},~ \tanh(W_a\circ \a{t}+\bm{b}_a)), & t>0.
        \end{cases}
\end{equation}
The per-step keypoint offsets output $\dx{t}$ (normalized into $[-1, 1]$) is then predicted based on the hidden state $\h{t}$ as $\dx{t} = \tanh(W_{\triangle x}\circ\h{t}+\bm{b}_{\triangle x})$.

\myparagraph{Learning.}
We sample all possible length-$T$ trajectories from the training dataset. Each trajectory $\tau$ is annotated with $\{\xGT{0:T}\}$ and $\{\a{1:T}\}$. The loss function is defined as the sum of the mean squared error (MSE) of both the 2D keypoint coordinates and the 2D keypoint offsets:
\begin{equation}
\centerline{$\ell = \sum_{\tau \in \D}\sum_{t=1}^T (\|\x{t}-\xGT{t}\|^2 + \|\dx{t}-\dxGT{t}\|^2)$.}
\end{equation}

\vspace{-10pt}
\subsection{Goal-Conditioned Policy} 
\vspace{-2pt}
\label{sec:policy}

We can now learn a policy that achieves the given task under the dynamics enforced by the learned dynamic model $\phi$. We formulate a Markov Decision Process and solve for a goal-conditioned policy, $\pi(\a{t} | \s{t}, \bm{g})$. In addition to the current state $\s{t}$, the policy also takes as input a goal vector $\bm{g}$ defined by the task. For example, the reaching task would define the goal vector as $\g{t} = [\x{t}_R - \bar{\bm{x}}_R, \x{t}_L - \bar{\bm{x}}_L]$, where $\bar{\bm{x}}_R$ and $\bar{\bm{x}}_L$ are the 2D target positions for the right wrist and the left wrist to reach, respectively. 

\myparagraph{Reward function.}
The goal of reinforcement learning is to solve for a policy that maximizes the long-term reward. Our reward function consists of the following four terms:
\vspace{-5pt}
\begin{equation}
    r = w_{\text{task}} r_{\text{task}} + w_{\text{upright}} r_{\text{upright}} + w_{\text{ctrl}} r_{\text{ctrl}} + w_{\text{alive}} r_{\text{alive}},
\end{equation}
where $w$ are weights. Among the four terms, $r_{\text{task}}$ encourages the policy to minimize the deviation from the goal of the task. For example, a reaching task would be defined as 
\vspace{-5pt}
\begin{equation}
    r_{\text{goal}}^{t} = -\left(\|\x{t+1}_R - \g{t+1}_R\|+\|\x{t+1}_L - \g{t+1}_L\|\right),
\end{equation}
$r_\text{upright}$ encourages the agent to maintain an upright torso by penalizing the positions of the neck and the head being lower than those in the default pose:
\vspace{-2pt}
\begin{equation}
    r_{\text{upright}}^{t} = -(\max(0, \x{t+1}_{\text{neck-y}}-h_{\text{neck}}) + \max(0,\x{t+1}_{\text{head-y}}-h_{\text{head}})),
\end{equation}
where $h_{\text{neck}}$ and $h_{\text{head}}$ are the height of the neck and of the head in the default pose. $r_{\text{ctrl}}$ encourages minimal action which leads to smoother motion: $r_{\text{ctrl}}^{(t)} = \|\a{t}\|$. Finally, $r_{\text{alive}}$ penalizes the policy from stepping out of the image boundary or falling down:
\vspace{-5pt}
\begin{multline}
        r_{\text{alive}} = \begin{cases} 1 , &  \text{\parbox{12em}{\small $x^{(t)}$ in image boundary and the height of neck and head are above a fixed threshold;}} \\ 0, & \text{\small otherwise}. \end{cases} 
\end{multline}
         
\label{eq:ralive}

\myparagraph{Training policy.}
We use Proximal Policy Optimization (PPO) \cite{schulman2017proximal} to train an MLP policy $\pi_\theta$ with a fixed horizon $T=50$. At the beginning of each episode, the initial 2D pose and the targets are independently randomly sampled from the training dataset we used to train the dynamic model. At each time step $t$, we clip the current 2D pose $\x{t}$ into a valid range. This range is determined by the boundary of 2D poses in $\mathcal{D}$. The episode is terminated whenever the policy reaches an invalid state that makes $r_{\text{alive}}=0$.

\myparagraph{Testing policy.}
Once the policy is trained, we can use it to generate 2D pose sequences that reach or track arbitrary targets/target trajectories while responding to the forceful perturbation applied by the user. The user-specified perturbation is interpreted as a torque applied on specific joints in a time interval. The torque will then be added to the output of the policy model.

\vspace{-4pt}
\subsection{Photorealistic Rendering}
\label{sec:render}
\vspace{-2pt}

    Starting from the input image $\I{0}$, we first extract the initial 2D pose $\x{0}$ with the state-of-the-art 2D pose detector~\cite{cao2019openpose}.
    We then use the policy $\pi$ and the dynamic model $\phi$ to produce a sequence of 2D poses $\{\x{1:T}\}$ for a given control task.
    A photorealistic video can then be synthesized frame-by-frame from the 2D poses:
        $\I{t} = \psi(\x{t})$ for $t = 1:T$,
    where $\psi$ is a generator network that maps the 2D pose to the corresponding image of the source person.\\
    For best photorealism, we adopt the method proposed in Everybody Dance Now~\cite{chan2019everybody} to construct $\psi$, and train it
    with an additional video where the source person performs random poses\footnote{Note that it is possible to omit this additional video input, if we do not care too much about the quality of the generated video. In that case, a pose transfer method (e.g. \cite{balakrishnan2018synthesizing}) can be used. This would require fewer inputs but also produce more artifacts in the generated video.}.  Our motion synthesis module is applicable to other pose-to-image generators, such as ADGAN~\cite{men2020controllable}. Although Everybody Dance Now~\cite{chan2019everybody} produces more temporally coherent videos, ADGAN is capable of generating results with various human subjects from the DeepFashion Dataset~\cite{liu2016deepfashion}. More results can be found in \secref{exp}. Note that the video training data can be completely independent of the 3D poses in the mocap training set. In this paper, we directly use mocap data from the public mocap database, Human 3.6M dataset~\cite{ionescu2013human3}, and record our own video data. \\

%% file: 4_experiments.tex
\begin{table*}[t]
\begin{subtable}[b]{0.49\textwidth}
\centering
\setlength{\tabcolsep}{3pt}
\small
    \begin{tabular}{lcc}
    \toprule
        Methods & realism & precision\\
        \midrule
        \model (ours) \vs Ground truth & 28\% & 36\% \\
        \model (ours) \vs Hao~\etal~\cite{Hao_2018_CVPR} & 88\% & 79\% \\
        \model (ours) \vs 3D simulation & 60\% & 61\% \\
    \bottomrule
    \end{tabular}
    \caption{User assessment on tracking and reaching task}
\end{subtable}
\hfill
\begin{subtable}[b]{0.49\textwidth}
\centering
\setlength{\tabcolsep}{3pt}
\small
    \begin{tabular}{lccc}
    \toprule
        Methods & realism & precision & perturbation\\
        \midrule
        \model (ours) \vs Hao~\etal~\cite{Hao_2018_CVPR} & 85\% & 74\% & 75\% \\
        \model (ours) \vs 3D simulation & 65\% & 74\% & 64\% \\
    \bottomrule
    \end{tabular}
    \caption{User assessment on tracking task with perturbation}
\end{subtable}
\vspace{-10pt} 
\caption{The results of the user study. The matrices include the percentage (\%) of users that prefer our method over the methods indicated in the first column. The following columns are the user assessment of video realism, tracking precision, and the ability to recover from a perturbed force, respectively. (a) User assessment of tracking and reaching videos. (b) User assessment of tracking videos that have an external force added on the pelvis during the movements.
}
\label{tab:userstudy}
\vspace{-10pt}
\end{table*}

\vspace{-16pt}
\section{Experiments}
\label{sec:exp}

This section presents the experiment setup, compares our method to baselines via quantitative metrics and user studies, and provides results for ablation studies and applications. 

\vspace{-4pt}
\subsection{Experimental Setup} 
\vspace{-2pt}
\label{sec:expsetup}

\paragraph{Motion synthesis.}
For training the dynamic model and the control policy, we employ a 3D mocap dataset from the Human 3.6M dataset~\cite{ionescu2013human3}, which is publicly available on requests. We currently focus on upper body motions; therefore, we only include action classes with minimal lower body movements:  \textit{Directions, Discussion, Greeting, Posing, TakingPhoto}. For all experiments in the paper, the motion is synthesized for $T=50$ steps, at 50 FPS. After the motion is synthesized, we apply an average smoothing with window size 11 to improve the visual quality.
We use a perspective camera with predefined intrinsics to obtain the 2D keypoint coordinates in the front view.

\myparagraph{Photorealistic rendering.} 
For training the generator of Everybody Dance Now~\cite{chan2019everybody}, the target person is required to perform random upper body movements in front of a camera that is roughly oriented in the front view. In practice, we found that 6--8 minutes of recording is enough for a 30 FPS camera. At the beginning of the recording, the person is expected to start with a T-pose, so that a 4-DoF linear transformation could be computed to normalize the 2D poses in the photo view into the same pose space as the motion synthesis module. The videos are synthesized at resolution $256\times 256$, 50 FPS. For comparisons, we use \cite{chan2019everybody} as the generator, while using ADGAN~\cite{men2020controllable} pre-trained on DeepFashion for some applications.

\myparagraph{Baseline.}
We consider two baselines. First, \emph{Hao~\etal~\cite{Hao_2018_CVPR}} propose a video generation model that allows detailed control over the motion of the generated video. Given an image and sparse flow vectors specified by users, the model can generate a dense flow map that warps the input frame along flow vectors. We use the ground truth trajectories as the input flow vectors to their model, and compare their synthesized videos with ours. We also consider a \emph{3D simulation} baseline. Given an input image of a person and user-specified 2D target points, we first predict the 3D COCO joints of the person in the input image using HMR \cite{kanazawa2018end}. We then run inverse kinematics (IK) to have a 3D humanoid mimic the initial pose according to the predicted body joints. In the meantime, we map the user-specified 2D target points into the 3D space. We fix the depth of the 3D target points to the initial depth of human wrists. With the above settings, we use IK to have the 3D humanoid track the 3D target points. Finally, we reproject the computed 3D joints back to 2D and use the photorealistic render to synthesize a video. 

\myparagraph{Evaluation metrics.}
We conduct a human preference study on Amazon Mechanical Turk to compare the perceptual quality of our results and the baseline methods. Also, we compare the quality of the synthesized videos frame by frame using two metrics: PSNR and LPIPS~\cite{Zhang_2018_CVPR}. The two metrics help estimate the similarity between the ground truth and the generated images. However, we find PSNR could deviate from human preference because of the diversity of plausible human motions.

\subsection{Results}

\label{sec:quantitative}
\noindent For quantitative comparisons, we consider three tasks:
\begin{itemize}
    \myitem \emph{Tracking}: Given an RGB image of a person and the desired trajectories of the left and right wrists, generate a video of the person tracking the trajectories;
    \myitem \emph{Reaching}: Similar to tracking, but fixed target locations for wrists are provided instead of full trajectories; 
    \myitem \emph{Perturbation recovery}: In this scenario, the person is perturbed suddenly by an impulse torque (\eg, pushing the person to the left) while following the policy to achieve the goal. The expected output is a video of the person trying to recover from the perturbation and complete the original task. For Hao~\etal~\cite{Hao_2018_CVPR}, we put an extra flow vector on the person's pelvis to represent an impulse torque.
    For 3D simulation, we use inverse dynamics to calculate the internal force of the 3D humanoid and add an external force on the upper torso within a time interval.
\end{itemize}

For each pair of the RGB image and trajectories, we generate three videos using our method and the two baseline methods and compare them using the user study and evaluation metrics.

\myparagraph{Human preference study.} We conduct the user study using MTurk. We communicate with participants and obtain their consent to use their responses. For each example, the study presents to the participants a pair of videos, generated by our model and a baseline, superimposed with the target points/trajectories. Participants are asked to answer the following two questions: (1) Which video looks more realistic? (2) Which video follows the target points/trajectories better? There is no time limit for answering questions. For tracking and reaching, we take 40 videos of a person doing some upper body movement as ground truth, and extract the 2D wrist keypoint coordinates as the target trajectories input to both our and baseline methods. 
Furthermore, we also perform pairwise comparisons between the ground truth and our results.
\tabref{userstudy}(a) shows the results in preference matrices among 100--200 answers for each question. In terms of realism, our \model system outperforms the baselines, and 28\% of the participants even think our results are more realistic than the ground truth. Note that a perfect method is expected to confuse people 50\% of the time, so a 28\% preference rate is encouraging.

\begin{figure}[t]
    \small
    \begin{center}
    \imgrid{55pt}{ccccc}{
       \rotatebox{90}{\hspace{5pt}Hao~\etal \cite{Hao_2018_CVPR}} &
       \im{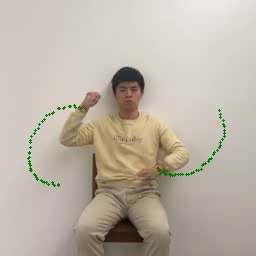} & 
       \im{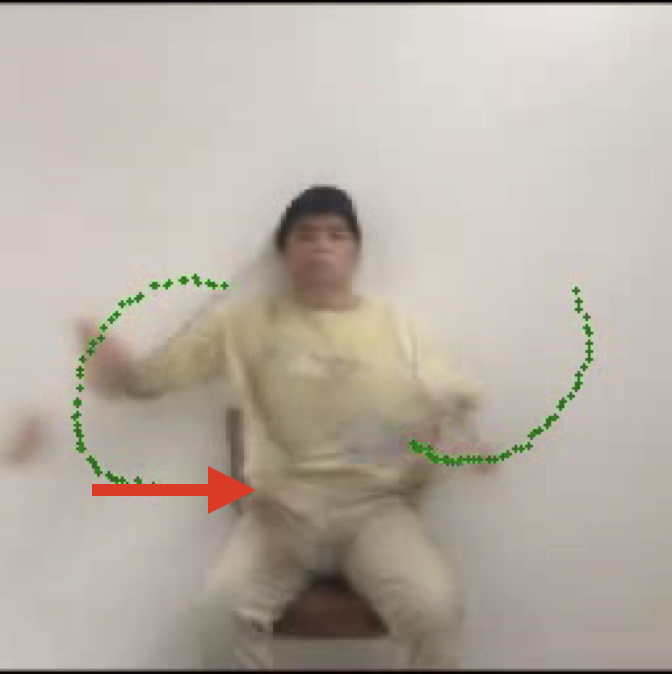} &
       \im{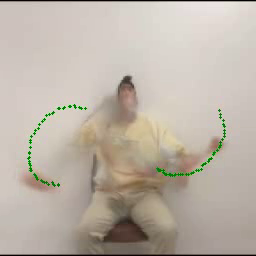} &
       \im{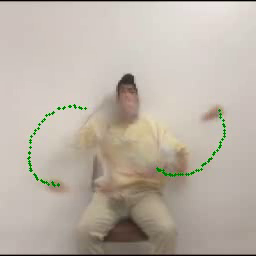}
       \\
       \rotatebox{90}{3D simulation} &
       \im{assets/qual_eval/perturb/Ours/000000.png} & 
       \im{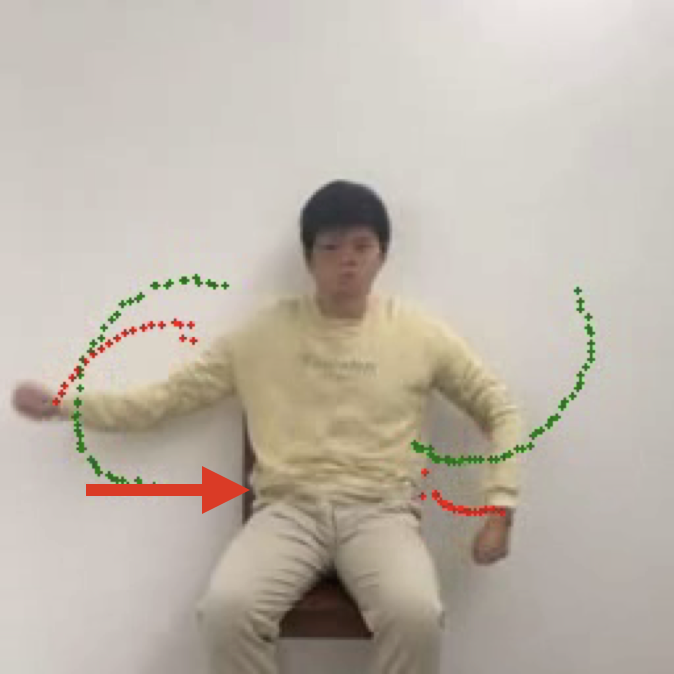} &
       \im{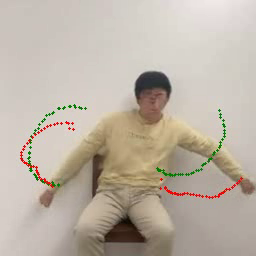} &
       \im{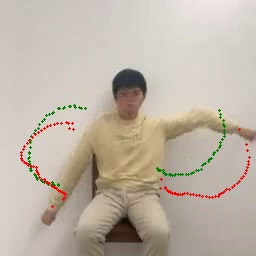}
       \\
       \rotatebox{90}{\hspace{18pt}Ours} &
       \im{assets/qual_eval/perturb/Ours/000000.png} & 
       \im{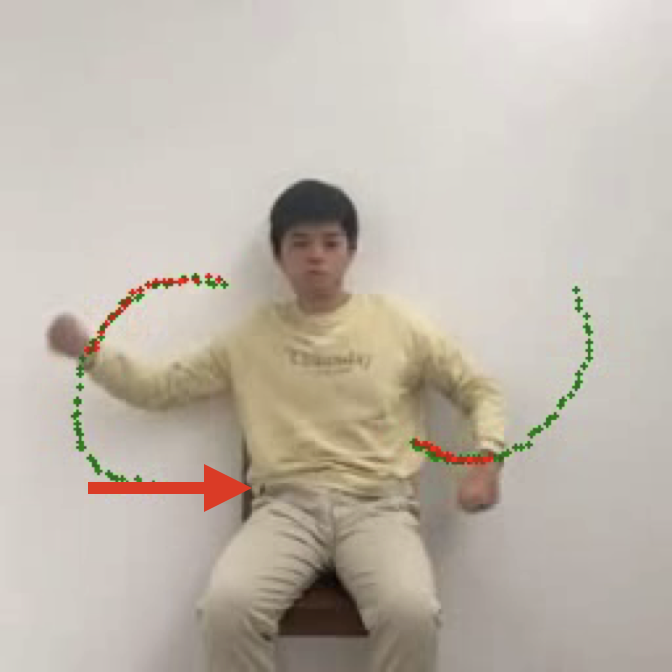} &
       \im{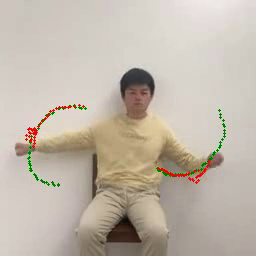} &
       \im{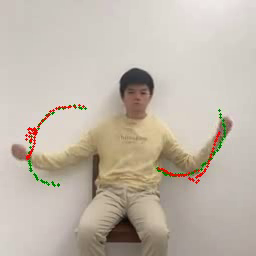}
       \\
       \rotatebox{90}{\hspace{8pt}SfV \cite{peng2018sfv}$^\ddagger$} &
       \im{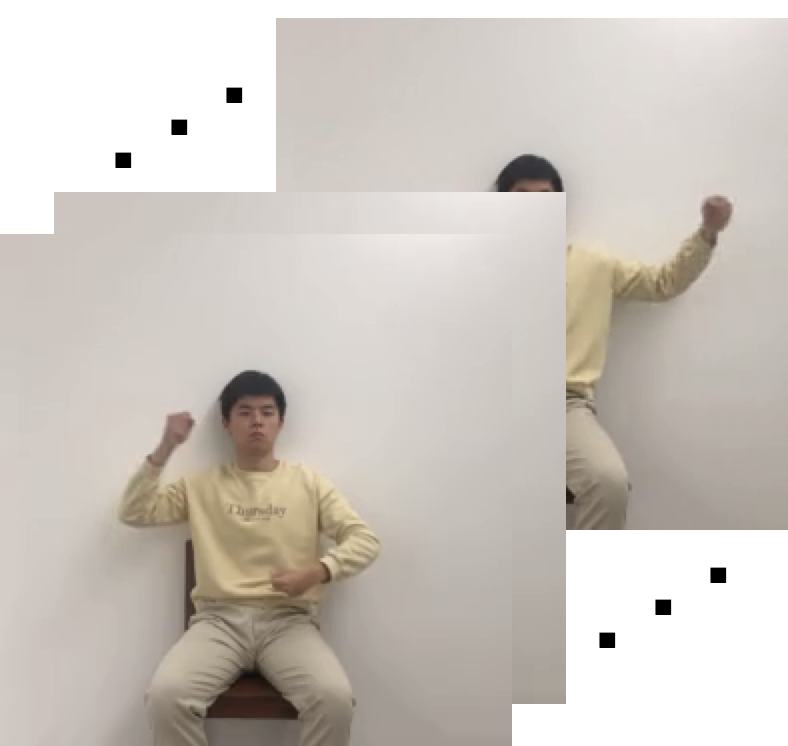} &
       \im{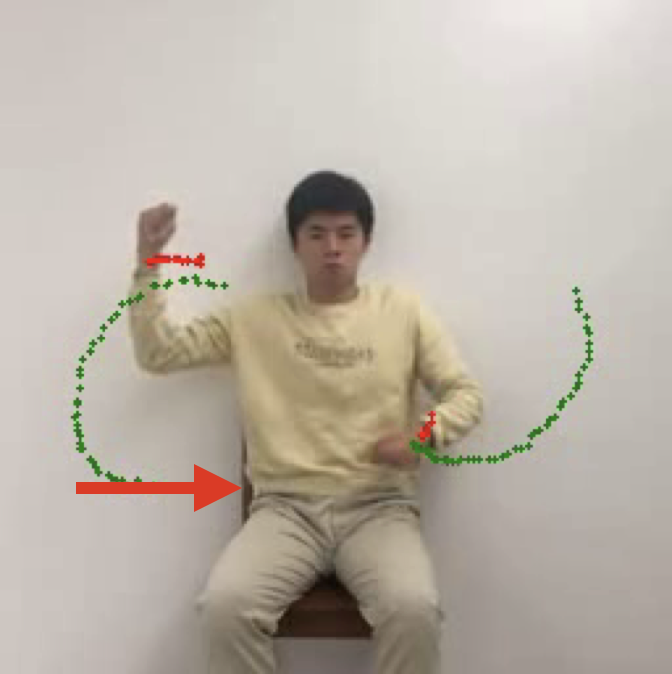} &
       \im{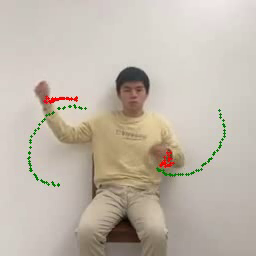} &
       \im{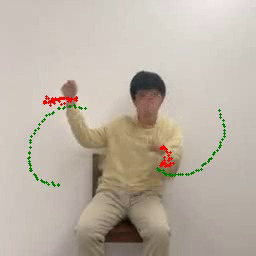} 
       \\
       \vspace{3pt}
       \\
        &
       Input & 
       $t = 20$ &
       $t = 40$ &
       $t = 50$ 
    }
    \end{center}
    \vspace{-15pt}
    \caption{The qualitative comparison with other methods on the perturbation task. Our method can recover from the perturbation better and keep tracking the trajectories more precisely. $^\ddagger$SfV \cite{peng2018sfv} requires the whole ground truth video as the input for training. See \secref{sfv} for more details.}
    \label{fig:qual_eval_ptb}
    \vspace{-15pt}
\end{figure}

For the perturbation recovery task, there is an additional question: In which video, does the person better recover from the sudden impulse torque? 
Note that there is no ground truth for the perturbation task. \tabref{userstudy}(b) reports the user preference of our method against the baselines.
Our method outperforms the baselines in all aspects. It is not surprising because our method leverages knowledge about human dynamics while \cite{Hao_2018_CVPR} attempts to model generic videos. Although 3D simulation utilizes the physics engine to generate reasonable results, the reconstructed initial pose contains inevitable errors. With an erroneous initial pose, the 3D simulation eventually generates a series of poses with more errors. Visually, we can observe misalignments of initial poses and wrists trajectories in \figref{qual_eval_ptb}. Besides, \cite{chan2019everybody} is susceptible to body misalignment, so the generated person by the 3D baseline is not as sharp as our method.

\myparagraph{Frame by frame evaluation.} 
In order to conduct a quantitative evaluation, we use a set of ten videos as the ground truth and evaluate the fidelity of the generated videos. \tabref{baselinemetrics} reports the average LPIPS~\cite{zhang2018unreasonable}, FID~\cite{heusel2017gans}, and FVD~\cite{unterthiner2018towards} scores of all methods. LPIPS and FID generally capture the similarity between two images, and FVD takes the temporal coherence of the generative videos into consideration. Our method outperforms the baselines in all metrics, and this is in line with the results of the user study.

\begin{table}[t]
    \small
    \centering
    \setlength{\tabcolsep}{3pt}
    \begin{tabular}{lccc}
        \toprule
        & LPIPS $\downarrow$ & FVD $\downarrow$ & FID $\downarrow$\\
        \midrule
        Hao~\etal~\cite{Hao_2018_CVPR} & 0.134 & 2197.59 & 330.96 \\
        3D Simulation & 0.164 & 1092.60 & 148.51 \\
        \model (ours) & {\bf 0.121} & {\bf 947.23} & {\bf 130.07} \\
        \bottomrule
    \end{tabular}
    \vspace{-3pt}
    \caption{Evaluation of our method and the baselines.}
    \label{tab:baselinemetrics}
\end{table}

\begin{table}[t]
    \centering\small
    \begin{tabular}{lcccc}
    \toprule
        \multirow{2}{*}{Methods} & \multicolumn{2}{c}{w/o perturbation} & \multicolumn{2}{c}{w/ perturbation} \\
        \cmidrule(lr){2-3}\cmidrule(lr){4-5}
        & $\mathcal{L}_1$ & $\mathcal{L}_2$ & $\mathcal{L}_1$ & $\mathcal{L}_2$ \\
        \midrule
        SfV \cite{peng2018sfv} & 26.41 & 20.66 & 31.73 & 25.10 \\
        3D simulation & 21.35 & 19.08 & 27.93 & 22.13 \\
        \model (ours) & \textbf{18.79} & \textbf{15.05} & \textbf{23.47} & \textbf{18.91} \\
    \bottomrule
    \end{tabular}
    \vspace{-5pt}
    \caption{Comparisons with the physical simulation methods. We compare the accuracy of the keypoints generated by different methods for the scenarios without and with perturbation. 
    }
    \label{tab:SfVtable}
    \vspace{-2pt}
\end{table}

\begin{figure}[t]
    \centering
    \includegraphics[width=\linewidth]{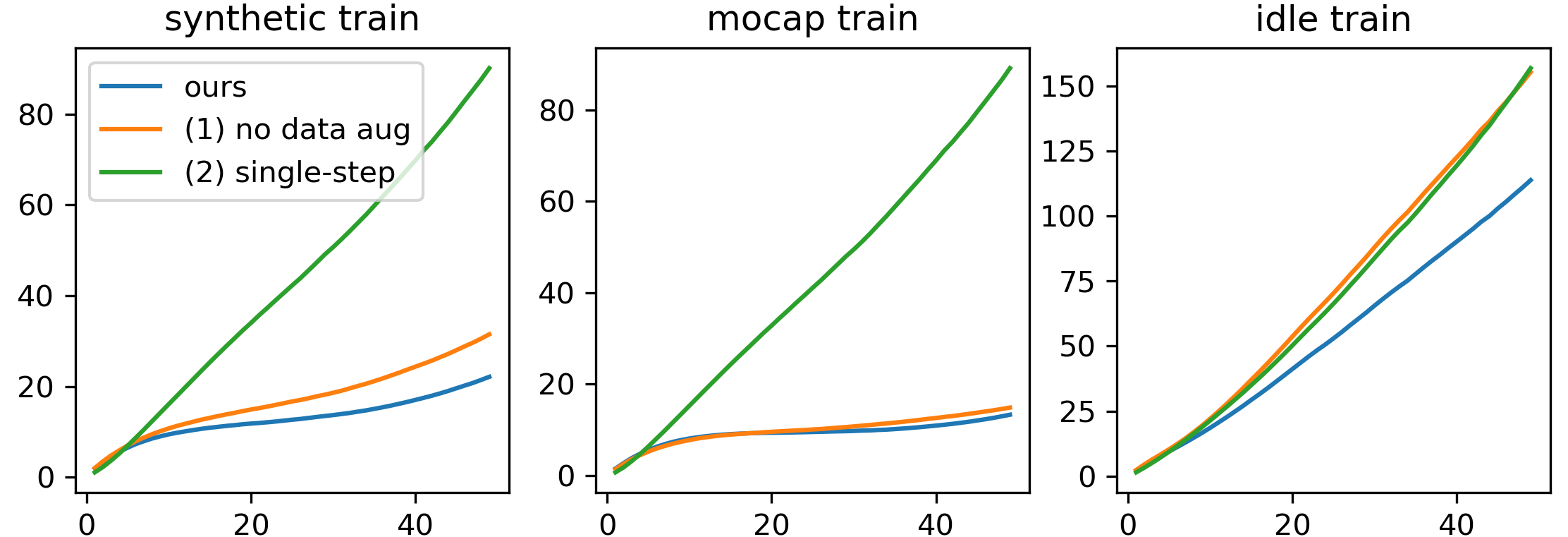}
    \vspace{-10pt}    
    \caption{Evaluating the dynamic model against two ablations: (1) no data augmentation and (2) single-step prediction. $x$-axis: step, $y$-axis: per-step average keypoint $\mathcal{L}_2$ error. \emph{Best viewed in color.}}
    \vspace{-10pt}    
    \label{fig:dyn-error}
\end{figure}

\begin{figure}[t]
    \small
    \imgrid{75pt}{ccc}{
       \im{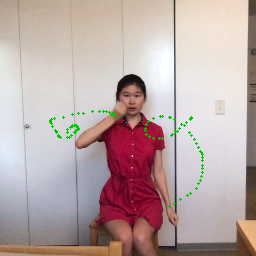} & 
       \im{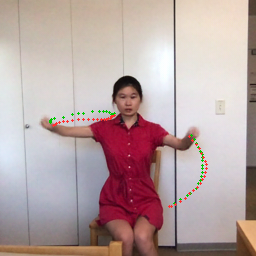} &
       \im{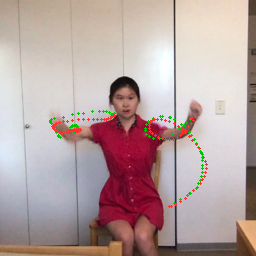} \\
       \im{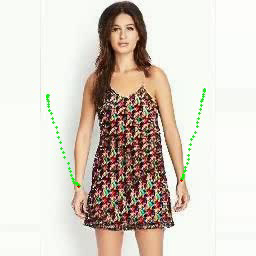} & 
       \im{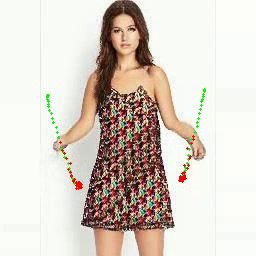} &
       \im{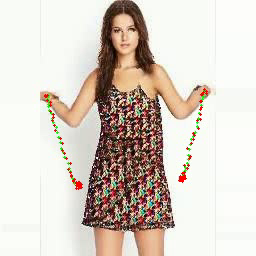} 
       \\
       \vspace{3pt}
       \\
       Input &
       $t = 25$ &
       $t = 50$
    } 
    \vspace{-3pt}
    \caption{Results of our approach tracking two different user-drawn trajectories. Leftmost: input images and the two input wrist trajectories (\textcolor{darkgreen}{green}); The rest columns show the trajectories of the synthesized wrist positions (\textcolor{darkred}{red}) overlaid on top of the synthesized frames. Top: video generated by Chan~\etal~\cite{chan2019everybody}; Bottom: video generated by Men~\etal~\cite{men2020controllable}.  \emph{Best viewed in color.} }
    \label{fig:tracking}
    \vspace{10pt}
    \small
    \imgrid{75pt}{ccc}{
       \im{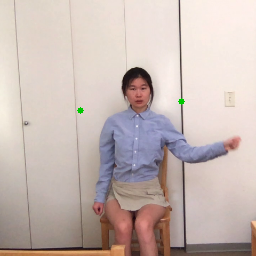} & 
       \im{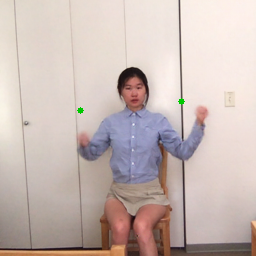} &
       \im{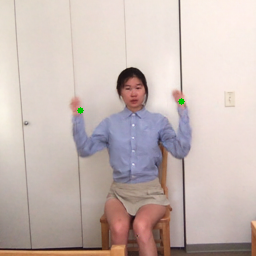} \\
       \im{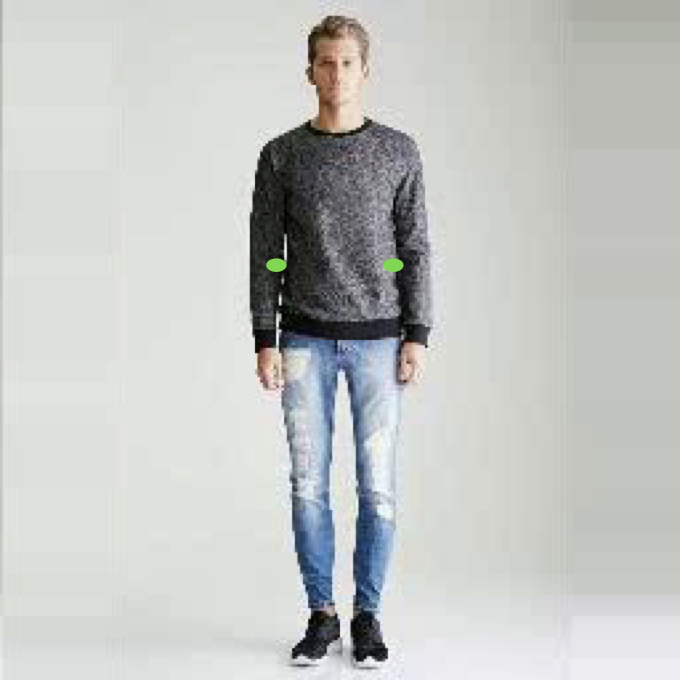} & 
       \im{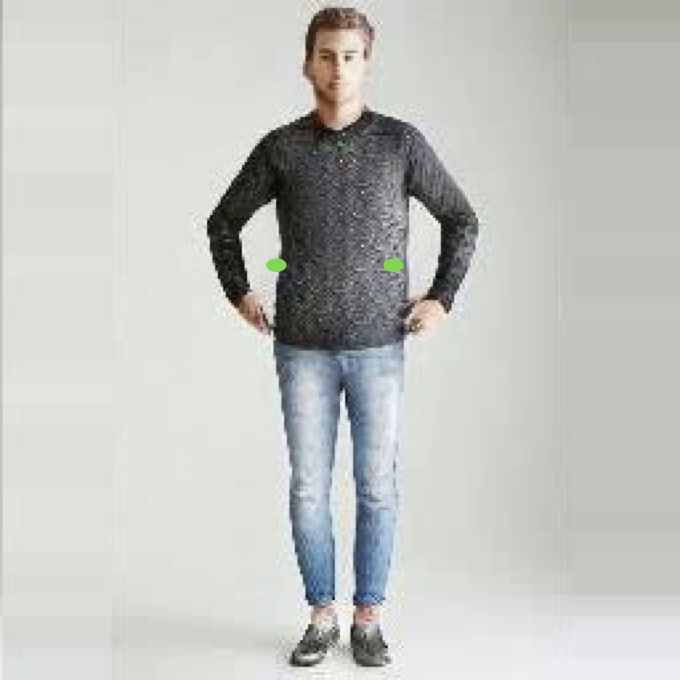} &
       \im{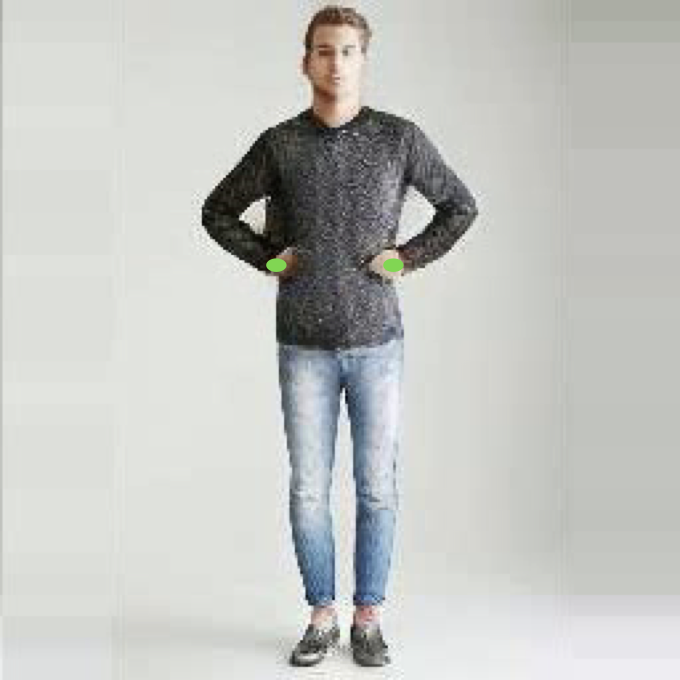} 
       \\
       \vspace{3pt}
       \\
       Input & 
       $t = 10$ & 
       $t = 20$ 
       \\
    }
    \vspace{-3pt}    
    \caption{Results of our approach reaching two different target wrist positions specified by the user. In the input images, the target wrist positions are marked with \textcolor{darkgreen}{green circles}; We show the synthesized frames at $t=10$ and $t=20$, demonstrating how the subjects reach targets. Top: video generated by the Chan~\etal~\cite{chan2019everybody}; Bottom: video generated by Men~\etal~\cite{men2020controllable}. \emph{Best viewed in color.}}
    \label{fig:reaching}
\vspace{-10pt}
\end{figure}    

\begin{figure}[t]    
     \small
    \imgrid{75pt}{ccc}{
       \im{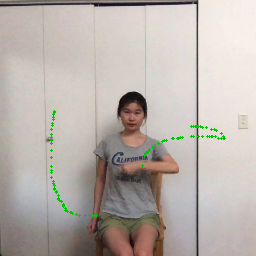} & 
       \im{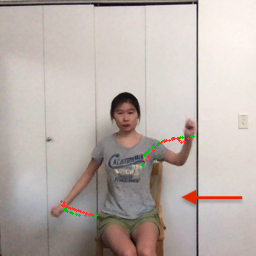} &
       \im{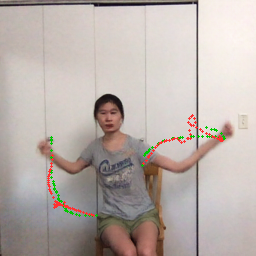} 
       \\
       \vspace{3pt}
       \\
       Input & 
       $t = 20$ &
       $t = 50$
       \\
    }
    \vspace{-3pt}    
    \caption{The person recovers from a push while tracking user-specified trajectories.The target trajectories are shown in (\textcolor{darkgreen}{green}), and the trajectories of the synthesized wrist positions are shown in (\textcolor{darkred}{red}). At $t=20$, an impulse torque is applied to the waist, which causes the person to rotate. \emph{Best viewed in color.}}
    \vspace{12pt}
    \label{fig:perturb}
        \small
    \imgrid{75pt}{ccc}{
       \im{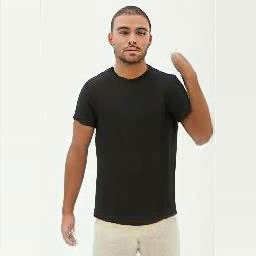} & 
       \im{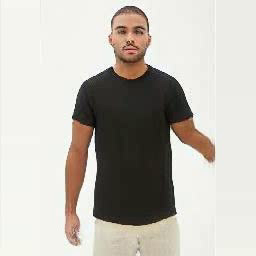} &
       \im{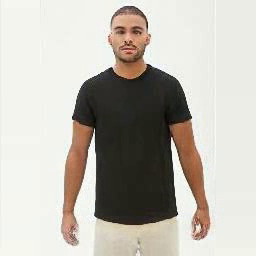} 
       \\
       \vspace{3pt}
       \\
       Input & 
       $t = 10$ &
       $t = 20$
       \\
    }
    \vspace{-3pt}      
    \caption{Force control. Arms dropping passively in response to gravity from $t =10$ to $t = 20$.}
    \label{fig:open-loop}
    \vspace{-15pt}
\end{figure}

\myparagraph{Other comparisons.}
\label{sec:sfv}
In addition to 3D simulation, we also compare our model with a more sophisticated 3D method, SfV~\cite{peng2018sfv}. Given a recorded human video, SfV enables a physically simulated 3D character to imitate the motions in the video. Note that we do not consider SfV as a baseline because SfV takes a whole ground truth video as input, and needs to train an RL model for each video to get the best performance, while our method only requires the first frame and wrist movements as input, and uses the same learned model for all videos. After inferring the motions, 3D simulation, SfV, and our method generate desired videos using the same photorealistic generator. Thus, we can compare the fidelity of the generated motions using the accuracy of the generated 2D keypoints. \tabref{SfVtable} reports the $\mathcal{L}_1$ and $\mathcal{L}_2$ distances between the 2D keypoints projected from Human 3.6M dataset~\cite{ionescu2013human3} and the ones generated by different methods. Our method outperforms the compared 3D methods in both metrics. Although SfV can mimic the target motion well, the tracking precision is worse than our method.

We also add an external force on the humanoids and see how the 3D methods resist the perturbation. We conduct HMMR~\cite{humanMotionKanazawa19} to predict the 3D motion of the ground truth video, and use a PD controller to track the ground truth motion with forward simulation. During perturbation, we add a force along the humanoid's pelvis. The PD controller will force the agent back to the trajectory after perturbation. \tabref{SfVtable} shows the results between the ground truth keypoints and the ones generated by different methods with perturbation. Our method generates more accurate poses because 3D simulation cannot recover from the perturbation well, and SfV fails to track the trajectory precisely after perturbation. \figref{qual_eval_ptb} compares them visually.

\vspace{-2pt}
\subsection{Ablation Study}
\vspace{-3pt}

We compare the dynamic model against two
ablations.
(1) \emph{no data augmentation}: the dynamic model is only trained on forces presented in the mocap dataset, without data augmentation; 
(2) \emph{single-step prediction}: the RNN is replaced with an MLP that takes in the 2D pose and 2D offset from the last step and predicts the 2D offset of the next step.

We evaluate the model on three setups.
(1) \emph{synthetic}: The augmented data as described in \secref{expsetup}. This tests how well the dynamic model generalizes when faced with highly perturbed torques, similar to the situation of policy training; 
(2) \emph{mocap}: The original mocap data, testing the performance on torques that lead to regular movements; 
(3) \emph{idle}: Starts from poses of the mocap data, but a zero torque is applied at each step. This tests the behavior when the person is only driven by gravity. The evaluation trajectories are drawn every 50 steps. 

\figref{dyn-error} shows the results. Our dynamic
model achieves the lowest prediction error on all three setups. It indicates that both expanding the coverage of force and including history information improve the robustness of the model.

\subsection{Applications}

\paragraph{Interactive photo.} 
The user can draw trajectories or specify  points on the image to be the targets for the wrists of the person. Although only the targets of the wrists are provided, our tracking policy is able to generate natural movements for the entire upper body. \figref{tracking} shows the result for the tracking task, and \figref{reaching} demonstrates the reaching task.

The user can also push the person in the image and see the person respond to the push and recover from the perturbation. \figref{perturb} shows examples of a person perturbed by a sudden leftward push around the waist area, while following the tracking policy. The impulse exerts from $t=20$ to $t=30$.

\myparagraph{Force control.} No goal is specified in this scenario, and thus the person is passively reacting to gravity and external forces. \figref{open-loop} shows that the person gradually drops her arms due to gravity. 
The generated movements are consistent with our physics intuition. More results can be found in the supplemental materials.

\begin{figure}[t]
    \small
   \imgrid{75pt}{ccc}{
       \im{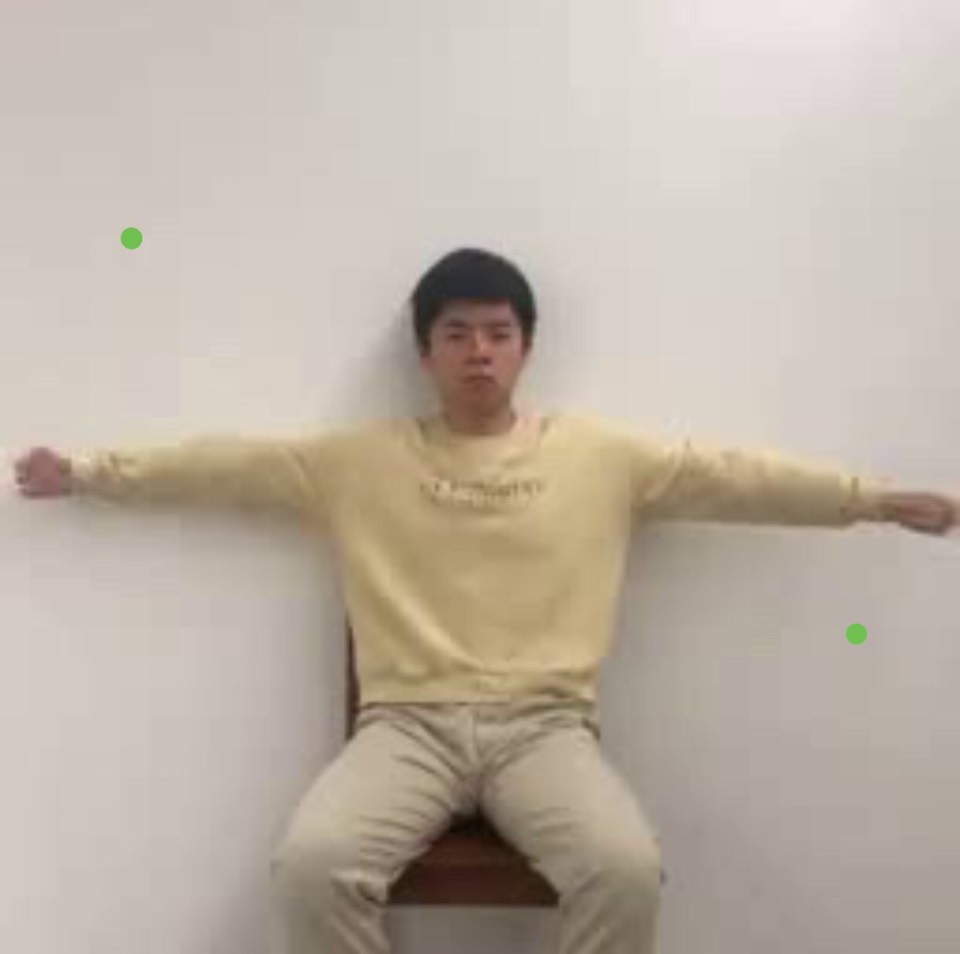} & 
       \im{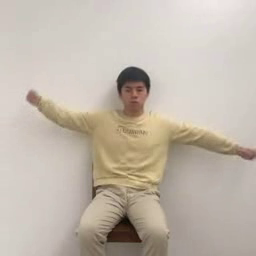} &
       \im{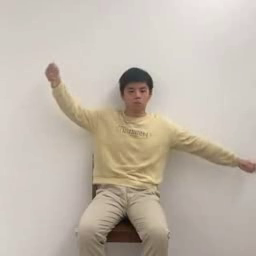} \\
       &
       \im{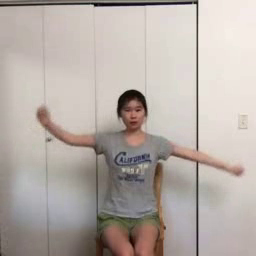} &
       \im{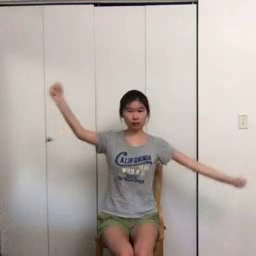} 
       \\
       \vspace{3pt}
       \\
       Input & 
       $t = 10$ &
       $t = 20$
       \\
    }
    \vspace{-3pt}      
    \caption{Retargeting. Top: An input image of the source subject and the results of retargeting. Bottom: Input images that specify the pose for retargeting.}
    \vspace{-5pt}      
    \label{fig:retargeting}
\end{figure}

\myparagraph{Retargeting.}  
We can also retarget the motion of one subject to other subjects with different appearances by aligning the source 2D pose sequence with the target generator's scale and pelvis position (\figref{retargeting}).

\begin{figure}[t]
    \small
   \imgrid{75pt}{ccc}{
       \im{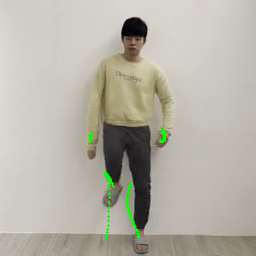} & 
       \im{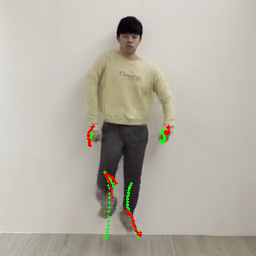} &
       \im{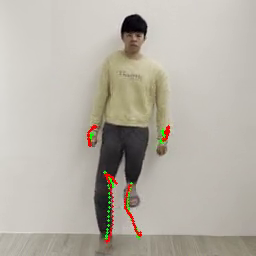}
       \\
       \\
       Input & 
       $t = 25$ & 
       $t = 50$
       \\
    }
    \vspace{-5pt}
    \caption{Results of the full-body tracking task. The leftmost image shows the input keypoints and four user-specified trajectories for wrists and ankles (\textcolor{darkgreen}{green}). Other images show the trajectories of the synthesized wrist and ankle positions (\textcolor{darkred}{red}) overlaid on top of the synthesized frames. \emph{Best viewed in color.}}
    \vspace{-10pt}
    \label{fig:full-body}
\end{figure}

\myparagraph{Full body control.}
We have tested our method on a small full-body dataset by fixing the pelvis and setting other joints free. The preliminary results are qualitatively similar to the upper-only model (\figref{full-body}); however, further tuning of reward function weight is required to achieve similar quantitative results. Further evaluation and optimization will remain for future research.

%% file: 5_discussion.tex
\vspace{-5pt}
\section{Discussion}
\label{sec:conclusion}
\vspace{-3pt}

We present a novel system, \model, that allows the user to generate photorealistic and physically plausible human animation by controlling a person in the image using intuitive 2D interaction. Experiments demonstrate that our framework is highly flexible and produces more compelling animation compared to the baseline. Importantly, we found that 2D keypoints already provide rich information and can be directly controlled effectively by 3D torques, removing the need to solve a challenging 3D pose reconstruction problem. Our method does not require 3D mocap data to be aligned with 2D images, largely simplifying the data acquisition process and enabling the composition of movements and appearances of different people. We also show that physics-based character animation can be an effective tool to synthesize training data that boost the performance of learning.

Our work has limitations. The dynamic model is not accurate when the input state-action pair is very different from those in the training data. We observe some bone stretching artifacts when the arms move to positions where training data are sparse. The data augmentation indeed helps, but it is somewhat task-dependent. When training a different category of motion, such as dancing, the policy might still visit the areas in state-action space in which the dynamic model performs poorly, resulting in a suboptimal dancing policy.
Training a policy that handles balance and physical interaction with the environment in the image space presents many interesting research challenges and applications.  

In our research, we collected videos of two human subjects and used data from the DeepFashion dataset~\cite{liu2016deepfashion}. Our research has been deemed IRB exempt by our institution, because we only used the human subject data as annotations, without studying the human subjects themselves. DeepFashion was released in 2016, widely used and cited over 1,000 times; the IRB approval situation of DeepFashion, including whether an IRB is needed, is unclear to us. We choose to use the dataset, because it was used by prior methods for state-of-the-art image generation results, which is critical for a fair comparison in our experiments.

Our research has many potential positive societal impacts, with future applications in the fashion industry, choreography, computer games, and personal and family entertainment with personalized, physically plausible avatars. It may also find uses in assistive technologies in the rehabilitation of disabled persons, and in automatic sign language generation for hearing impaired people. On the other hand, like all other visual content generation methods, our method might be exploited by malicious users to generate disinformation.
Thus, we urge users of our models to be aware of ethical and societal concerns and to apply them with good intent. We will use techniques such as watermarking to identify and label visual content generated by our system.